\documentclass[journal]{IEEEtran}

\usepackage{verbatim}
\usepackage{amsfonts}
\usepackage{amssymb}
\usepackage{stfloats}
\usepackage{cite}
\usepackage{graphicx}
\usepackage{psfrag}
\usepackage{amsmath}
\usepackage{array}
\usepackage{epstopdf}
\usepackage{authblk}
\usepackage{graphicx} 
\usepackage{amsthm} 
\usepackage{lipsum}
\usepackage{verbatim} 
\usepackage{authblk}
\usepackage{mathtools}
\usepackage{cuted}
\usepackage{booktabs}

\usepackage{amsmath}
\usepackage{mathrsfs}

\usepackage{algorithmic}

\usepackage{array}

\ifCLASSOPTIONcompsoc
\usepackage[caption=false,font=normalsize,labelfont=sf,textfont=sf]{subfig}
\else
\usepackage[caption=false,font=footnotesize]{subfig}
\fi

\usepackage{url}
\usepackage{graphicx,amsmath,amssymb,amsfonts}
\usepackage{algorithmic,algorithm}

\usepackage{hyperref}
\hypersetup{colorlinks=true, citecolor=blue}

\usepackage{setspace}

\makeatletter

\newcommand{\Rmnum}[1]{\expandafter\@slowromancap\romannumeral #1@}
\makeatother

\usepackage{framed} 

\usepackage{cancel}

\usepackage{booktabs}
\usepackage{tabularx,makecell,multirow}

\usepackage[table]{xcolor}
\usepackage{pifont} 

\usepackage{diagbox}

\begin{document}
	\bstctlcite{ref:BSTcontrol}

	\title{Federated Intelligence: When Large AI Models Meet Federated Fine-Tuning and Collaborative Reasoning at the Network Edge}
	
	\author{Wanli Ni, Haofeng Sun, Huiqing Ao, and Hui Tian
		
	\thanks{Wanli Ni is with the Department of Electronic Engineering, Tsinghua University, China (e-mail: niwanli@tsinghua.edu.cn).}
	\thanks{Haofeng Sun, Huiqing Ao, and Hui Tian are with the National Key Laboratory of Networking and Switching Technology, Beijing University of Posts and Telecommunications, Beijing 100876, China (email: sunhaofeng@bupt.edu.cn, hqao@bupt.edu.cn, tianhui@bupt.edu.cn).}
}
	
	\maketitle
	
	\begin{abstract}
	Large artificial intelligence (AI) models exhibit remarkable capabilities in various application scenarios, but deploying them at the network edge poses significant challenges due to issues such as data privacy, computational resources, and latency.
	In this paper, we explore federated fine-tuning and collaborative reasoning techniques to facilitate the implementation of large AI models in resource-constrained wireless networks.
	Firstly, promising applications of large AI models within specific domains are discussed.
	Subsequently, federated fine-tuning methods are proposed to adapt large AI models to specific tasks or environments at the network edge, effectively addressing the challenges associated with communication overhead and enhancing communication efficiency.
	These methodologies follow clustered, hierarchical, and asynchronous paradigms to effectively tackle privacy issues and eliminate data silos.
	Furthermore, to enhance operational efficiency and reduce latency, efficient frameworks for model collaborative reasoning are developed, which include decentralized horizontal collaboration, cloud-edge-end vertical collaboration, and multi-access collaboration.
	Next, simulation results demonstrate the effectiveness of our proposed methods in reducing the fine-tuning loss of large AI models across various downstream tasks.
	Finally, several open challenges and research opportunities are outlined.
	\end{abstract}

	\section{Introduction}
	The rapid evolution of artificial intelligence (AI) has led to the development of large AI models (e.g., ChatGPT and Llama) that capture more nuanced patterns and subtleties in data, thereby exhibiting remarkable capabilities in various application scenarios, especially for content generation \cite{Xu2024Unleashing, Quan2025Large}.
	However, pre-training, fine-tuning and deploying these large models in practical Internet of Things (IoT) networks pose significant challenges, particularly due to issues related to data privacy, computational resources, and latency \cite{Cheng2024Towards}.
	Existing research has explored various federated fine-tuning approaches to address the privacy concerns associated with centralized AI model training \cite{Sun2024Federated}.
	These approaches employ parameter-efficient training techniques, such as low-rank adaptation (LoRA) and prompt tuning, to minimize computational costs \cite{Chen2024The}.
	Furthermore, to mitigate reasoning latency and tackle the deployment challenges posed by large AI models on resource-constrained devices, researchers have proposed collaborative reasoning frameworks that enable the distribution of computational workloads for downstream tasks across multiple devices or servers \cite{Xu2024OnDevice}.
	
	In the era of large AI models, the integration of federated fine-tuning and collaborative reasoning offers significant benefits for achieving federated intelligence \cite{Zhang2024FederatedIntelligence}.
	Such a collective intelligence approach makes it well-suited to tackle complex AI tasks in distributed and resource-constrained environments \cite{Chen2024The}.
	By leveraging federated intelligence, large AI models distributed in real-world wireless networks can collectively form comprehensive perspectives and collaborative efforts. This is particularly effective in tackling complex multi-disciplinary problems that exceed the capabilities of a single AI model.
	Specifically, through the use of federated fine-tuning, large AI models can be trained on a broader and more diverse range of data without compromising privacy, ultimately resulting in more desirable and effective models \cite{Chen2024The}.
	Similarly, collaborative reasoning can help distribute the computational load of computing tasks, making large AI models more accessible and efficient among resource-constrained devices \cite{Cai2024Edge}.
	Overall, this integration holds the promise of unlocking the federated intelligence of large AI models within wireless networks, and offers a range of benefits, such as reduced communication and computation overheads, enhanced model personalization, and strengthened privacy protection \cite{Long2024The}.

	Although federated intelligence holds immense potential in future wireless networks, it also poses substantial challenges \cite{Cheng2024Towards}.
	On the one hand, fine-tuning large AI models directly on resource-constrained IoT devices is impractical due to their limited computing power and storage capacity.
	The authors of \cite{Sun2024Federated} proposed a federated LoRA scheme to enable efficient fine-tuning of large language models (LLM) over wireless networks, and utilized an over-the-air computation-based transmission scheme to achieve fast parameter aggregation.
	On the other hand, determining the optimal partitioning of a large AI model across multiple devices, or efficiently breaking down complex inference tasks into smaller, cooperative subtasks, can be difficult \cite{Lin2024SplitLoRA}.
	The authors of \cite{Ao2025FSL} proposed a semi-asynchronous split learning framework to address the challenges of device heterogeneity and waiting latency, where a Lyapunov-based online optimization algorithm was designed to dynamically adjusts the resource allocation and model partitioning based on real-time channel conditions.
	However, existing research lacks a systematic review of federated intelligence, particularly from the perspectives of federated fine-tuning and collaborative reasoning for large AI models in wireless networks.
	Against this background, the following contributions are made in this paper:
	\begin{itemize}
		\item
		\textbf{Federated Fine-Tuning:}
		We provide three federated fine-tuning schemes for training large AI models in wireless networks, each designed to address specific challenges.
		For example,
		1) \textit{Cluster federated fine-tuning} groups similar users into clusters, allowing for more targeted and efficient model updates, which is particularly effective in environments with heterogeneous data distributions.
		2) \textit{Hierarchical federated fine-tuning} introduces a hierarchical structure to the federated learning process that helps balance the trade-off between generalization and personalization.
		3) \textit{Asynchronous federated fine-tuning} allows devices to participate in the learning process at their own pace without requiring synchronization between all devices, which is critical for devices with different availability and connectivity.
		In summary, these methods not only improve the performance of large AI models on a wide range of downstream tasks but also ensure the protection of user privacy, making them suitable for model fine-tuning in various domains.
		
		\item
		\textbf{Collaborative Reasoning:}
		We offer a set of collaborative reasoning frameworks to optimize resource utilization and improve inference efficiency for complex tasks in distributed environments.
		These frameworks include: 
		1)~\textit{Decentralized horizontal collaboration} distributes inference tasks among multiple devices within a peer-to-peer network to boost resource utilization.
		2) \textit{Cloud-edge-end vertical collaboration} integrates the computational resources of cloud, edge, and end devices to minimize response time.
		3) \textit{Multi-access collaboration} enables devices to dynamically select the most appropriate communication pathway to ensure high reliability and low latency.
		Overall, these frameworks provide a robust and scalable solution for enhancing the reasoning performance of large AI models at the network edge.
	\end{itemize}

	The remainder of this paper is organized as follows.
	First of all, Section II discusses the applications of large AI models in specific domains, followed by Section III, which delves into federated fine-tuning methods to effectively preserve user privacy.
	Section IV explores collaborative reasoning frameworks for allocating the inference tasks across multiple devices or servers.
	Section V presents the simulation results, while Section VI discusses future research opportunities. Finally, Section VII concludes the paper.
	
	\begin{figure*}[t]
		\centering
		\includegraphics[width=6.8 in]{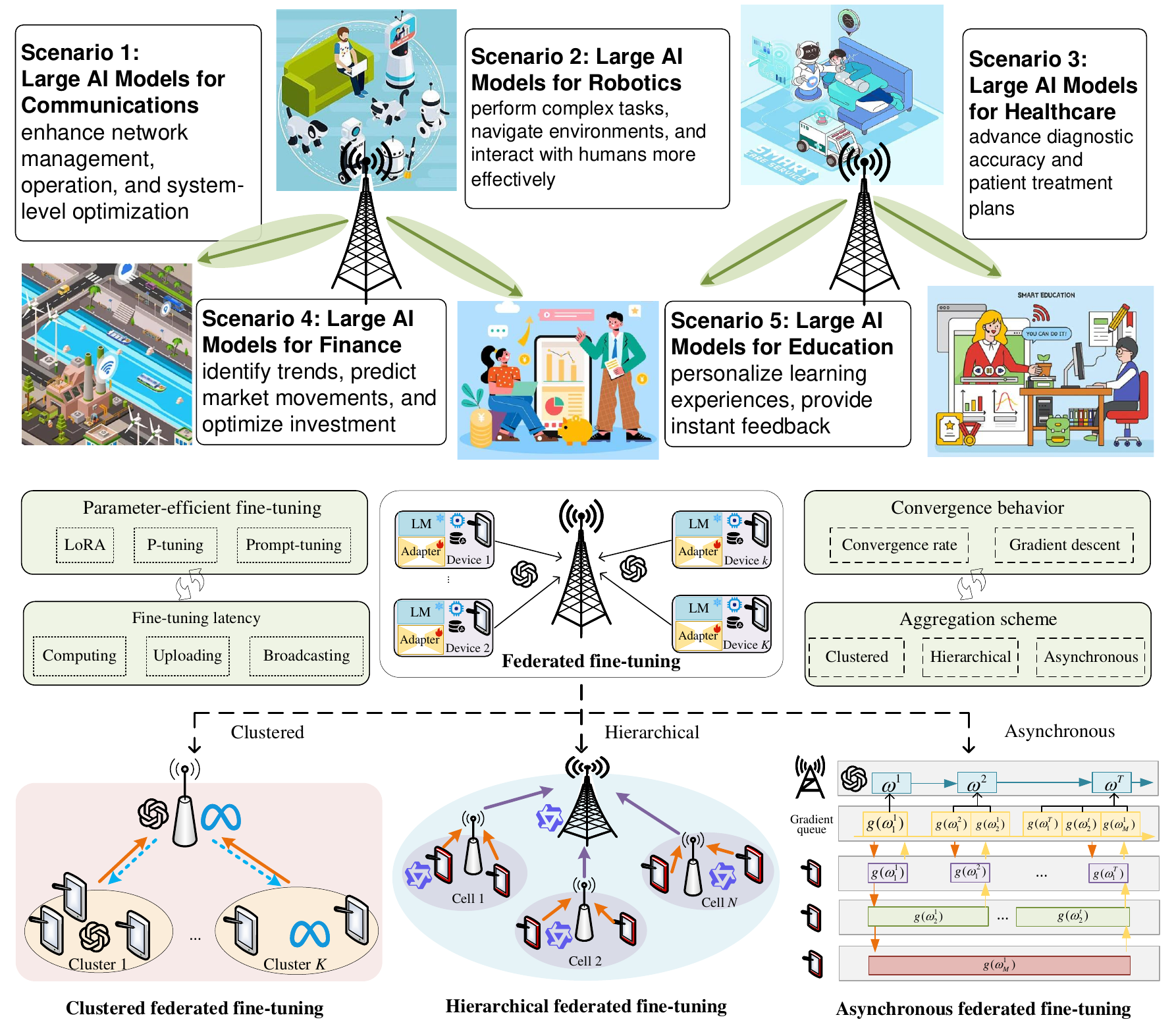}
		\caption{Diverse application scenarios of large AI models, and different federated learning methods for fine-tuning models in wireless networks.}
		\label{Fig1}
	\end{figure*}

	\section{Applications of Large AI Models in Specific Domains}
	Large AI models hold the promise of revolutionizing numerous industries by providing unprecedented insights and innovative solutions. Below is a detailed introduction to how these models are being utilized in the communications, robotics, healthcare, finance, and education domains.
	
	\textbf{Large AI Models for Communications:}	
	In communication systems, large AI models can be used to enhance network efficiency, optimize data transmission, and improve user experience.
	One key area where these models are making a significant impact is in network management, operation, and system-level optimization. By leveraging their advanced prediction capabilities, large AI models can predict and adapt to network traffic patterns in real-time, enabling more efficient resource allocation and reducing latency.	
	Furthermore, large AI models are also being utilized in physical layer design. By analyzing user behavior and network conditions, large models can identify the optimal coding and modulation strategies, leading to improved spectral efficiency.

	\textbf{Large AI Models for Robotics:}	
	In robotics, large AI models are instrumental in advancing the capabilities of autonomous systems. These models enable robots to understand and navigate complex environments with greater precision and adaptability. By processing and interpreting multi-modal sensory data from cameras, lidars, and other sensors, they allow robots to perform tasks such as object recognition, path planning, and obstacle avoidance with remarkable accuracy. Moreover, large AI models are being used to enhance the learning abilities of robots, enabling them to continuously improve their performance through experience and interaction with their surroundings. This is particularly important in fields like manufacturing, where robots need to adapt to varying production processes and product specifications.
	
	\textbf{Large AI Models for Healthcare:}	
	In healthcare, large AI models are transforming the way medical data is analyzed and utilized. These models can process vast amounts of clinical data, including patient records, imaging studies, and genomic sequences, to identify patterns and insights that may not be apparent to human analysts. This has led to significant advancements in disease diagnosis, treatment planning, and patient monitoring. For instance, large AI models have been used to develop early warning systems for critical conditions such as sepsis and heart failure, as well as to personalize treatment plans based on individual patient characteristics.
	
	\textbf{Large AI Models for Finance:}	
	In the financial industry, large AI models are revolutionizing risk management, fraud detection, and investment strategy. They can analyze historical financial data, market trends, and economic indicators to forecast future market movements and identify potential risks. This has enabled financial institutions to make more informed investment decisions and manage their portfolios with greater precision. Furthermore, large AI models are being used to detect and prevent fraud by analyzing transaction patterns and identifying anomalies that may indicate suspicious activity. 
	
	\textbf{Large AI Models for Education:}	
	In education, large AI models are transforming the way students learn and teachers teach. These models can personalize learning experiences by analyzing student data, such as test scores, assignment submissions, and engagement levels, to identify strengths, weaknesses, and areas for improvement. This information is then used to tailor educational content and resources to meet individual student needs. Additionally, large AI models are being used to develop intelligent tutoring systems that can provide real-time feedback and guidance to students as they work through problems and assignments.

	\section{Federated Fine-Tuning for Large AI Models}
	Owing to the issues such as privacy preservation, data heterogeneity, and scalability, it is crucial to investigate federated fine-tuning methods when deploying large AI models in resource-constrained IoT networks.
	As shown in Fig.~\ref{Fig1}, federated fine-tuning allows for the adaptation of pre-trained large models to specific tasks or domains without necessitating the centralization of the local raw data from end devices, thereby preserving user privacy. 
	In the following, three variations of federated fine-tuning are discussed.
	
	\subsection{Clustered Federated Fine-Tuning}	
	Based on the factors such as data similarity, computational capabilities, or communication latency, clustered federated fine-tuning first entails the formation of clusters comprising participating clients \cite{Feng2022Mobility}.
	Subsequently, each cluster independently proceeds with its own federated fine-tuning process, wherein model updates are aggregated within the cluster prior to potential sharing or further aggregation across clusters.
	This approach leverages the natural heterogeneity in federated environments, enabling more tailored model customization within clusters while maintaining privacy. 
	By reducing the variability in data distributions and computational resources within each cluster, clustered federated fine-tuning can accelerate the convergence of model training and improve the final model's performance on local tasks. 
	
	\subsection{Hierarchical Federated Fine-Tuning}	
	By introducing a hierarchical structure within the wireless network, hierarchical federated fine-tuning builds upon the clustered approach. 
	In this setup, clients are organized into multiple layers of hierarchy, wherein lower-level clients form clusters that subsequently function as clients in higher-level federations \cite{Xu2022Adaptive}.
	This multi-layered architecture facilitates the refinement of models at various levels of granularity. 
	For instance, models might first be fine-tuned within local clusters (representing small communities or organizations), and then these locally fine-tuned models can be aggregated and further fine-tuned at regional or global levels. 
	This hierarchical approach facilitates the balance between model generalization across broader users and customization to specific user groups. 
	It caters to varying levels of privacy concerns and regulatory requirements by enabling more precise control over data sharing and model updates.
	
	\subsection{Asynchronous Federated Fine-Tuning}	 
	In synchronous federated fine-tuning, the requirement for all clients to participate simultaneously in each training round may pose practical challenges due to variations in network conditions, client availability, and computational capabilities.
	Asynchronous federated fine-tuning circumvents these issues by allowing clients to update the model independently. 
	Clients can download the latest global model whenever they are available, perform local training, and upload their updates at their convenience. 
	The server then aggregates these model updates as they arrive, continuously refining the global model. 
	This flexibility accommodates real-world constraints and accelerates the overall training process by maximizing the utilization of available resources. 
	However, asynchronous updates may introduce stale gradients and non-deterministic behavior, necessitating robust aggregation strategies and convergence guarantees to ensure model quality \cite{Ao2025FSL}.

	\begin{figure*}[t]
		\centering
		\includegraphics[width=6.9 in]{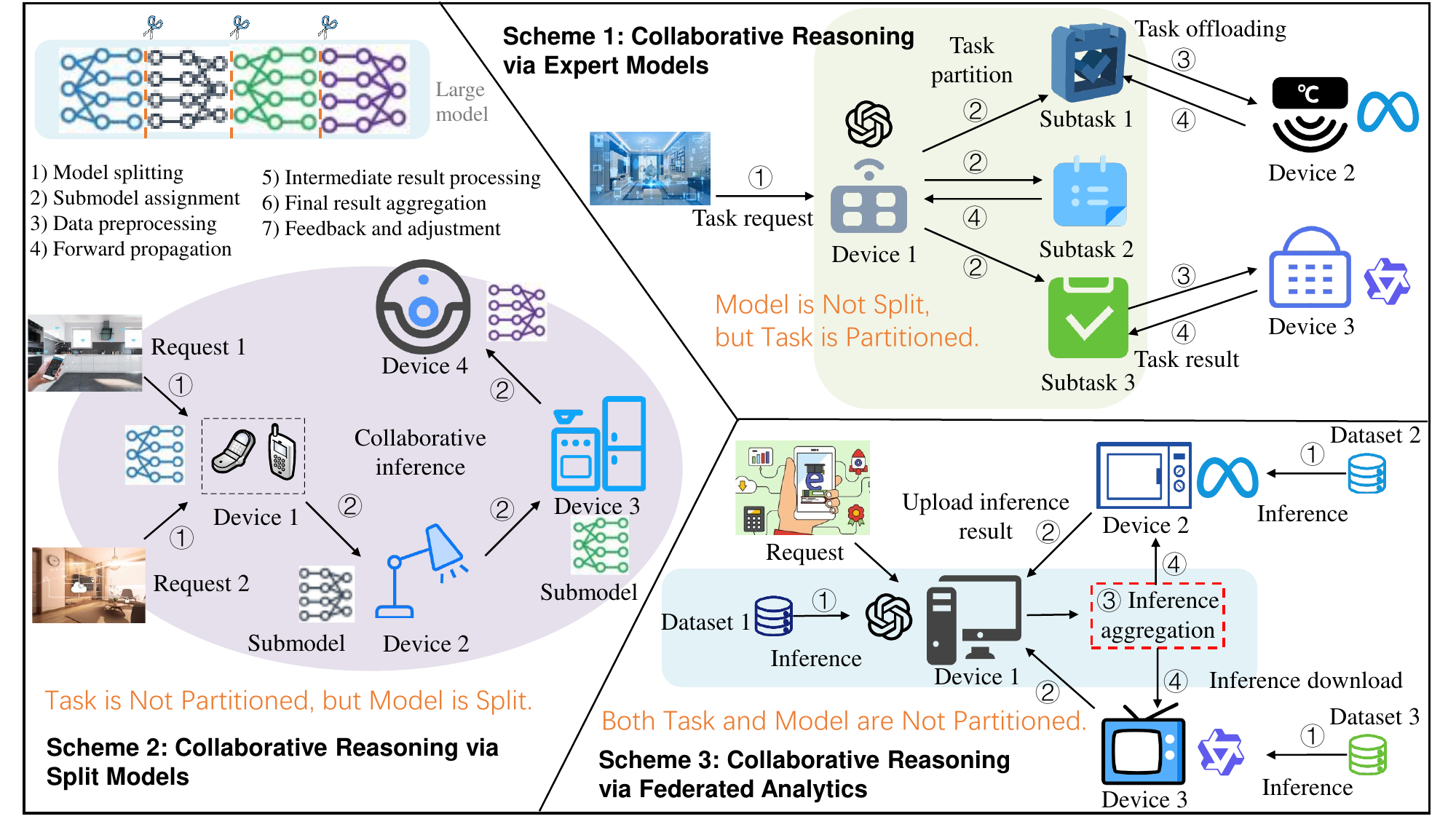}
		\caption{An illustration of the decentralized schemes for horizontal collaborative reasoning of large AI models at the network edge.}
		\label{Fig2}
	\end{figure*}

	\subsection{Convergence Analysis of Federated Fine-Tuning}	
	The examination of the convergence properties of federated fine-tuning methodologies is essential for ensuring their reliability and efficacy. This entails a comprehensive analysis of the convergence dynamics under diverse conditions, including heterogeneous data distributions, asynchronous updates, and constraints on communication resources. By establishing the upper bounds of convergence for federated fine-tuning, it becomes possible to enhance the model training process through the rational optimization of communication, computing, and aggregation strategies. 
	For instance, in our prior work \cite{Sun2024Federated}, we implemented a federated low-rank adaptation approach to fine-tune large models within wireless networks. Our findings indicated that the aggregation of low-rank adapters with a higher rank could improve fine-tuning performance, but it also significantly elevated the probability of encountering stragglers with weak computing power.

	\section{Collaborative Reasoning for Large AI Models}
	Despite the benefits of federated fine-tuning, deploying large AI models on resource-constrained devices still faces challenges such as high memory usage and slow inference speed.
	As models scale up, collaborative reasoning frameworks become crucial, as they not only enhance operational efficiency but also minimize response times for mobile devices.
	
	\subsection{Decentralized Horizontal Collaborative Reasoning}
	In an environment without a centralized node with powerful computing resources, decentralized collaborative reasoning paradigms are particularly well-suited for large AI models that require significant computational power and memory to perform complex inference tasks.
	As shown in Fig. \ref{Fig2}, we present the following three decentralized schemes for the horizontal collaborative reasoning.
	\begin{enumerate}
		\item
		\textbf{Collaborative Reasoning via Expert Models:} 
		In this scheme, each device is equipped with an expert model that is good at a certain class of tasks.
		When a computationally intensive task arrives at a particular device, it breaks down the task into several smaller, more manageable subtasks. These subtasks are then distributed to neighboring devices via device-to-device communications.
		Once the neighboring devices complete their assigned subtasks, they send the results back to the originating device.
		Note that in this scheme, the tasks are separable, but the model deployed on each device remains complete.
		However, expert model-based collaborative reasoning may entail considerable communication overhead and synchronization challenges due to the need to transmit subtasks and receive results among devices.
		
		\item 
		\textbf{Collaborative Reasoning via Split Models:}
		In split learning-based scheme, the model is segmented into several smaller submodels, each of which is deployed on a local IoT device \cite{Ao2025FSL}. These devices communicate with each other to coordinate the reasoning process, thereby guaranteeing the accuracy and consistency of the overall model's output. The division of the model can be customized according to a range of considerations, including the computational requirements of each component, the communication bandwidth available between nodes, and the privacy concerns related to the data being processed.
		Since submodels are distributed on different devices, the failure of a single device can compromise the entire reasoning process. Therefore, ensuring the robustness and resilience of the system is crucial.
		
		\item
		\textbf{Collaborative Reasoning via Federated Analytics:} 
		Federated analytics represents a cutting-edge approach for solving complex computational tasks by engaging multiple devices in a collective effort to perform result inference, without the need to share raw data \cite{Zhang2024FederatedIntelligence}. Within this scheme, neither the task nor the model is partitioned among the devices. Instead, each device conducts its computations locally, and the results are subsequently synthesized to produce a comprehensive global analysis. The aggregation process of federated analytics guarantees that the local insights from each device are integrated into the final outcome.
	\end{enumerate}
	
	\begin{figure}[t]
		\centering
		\includegraphics[width=3.5 in]{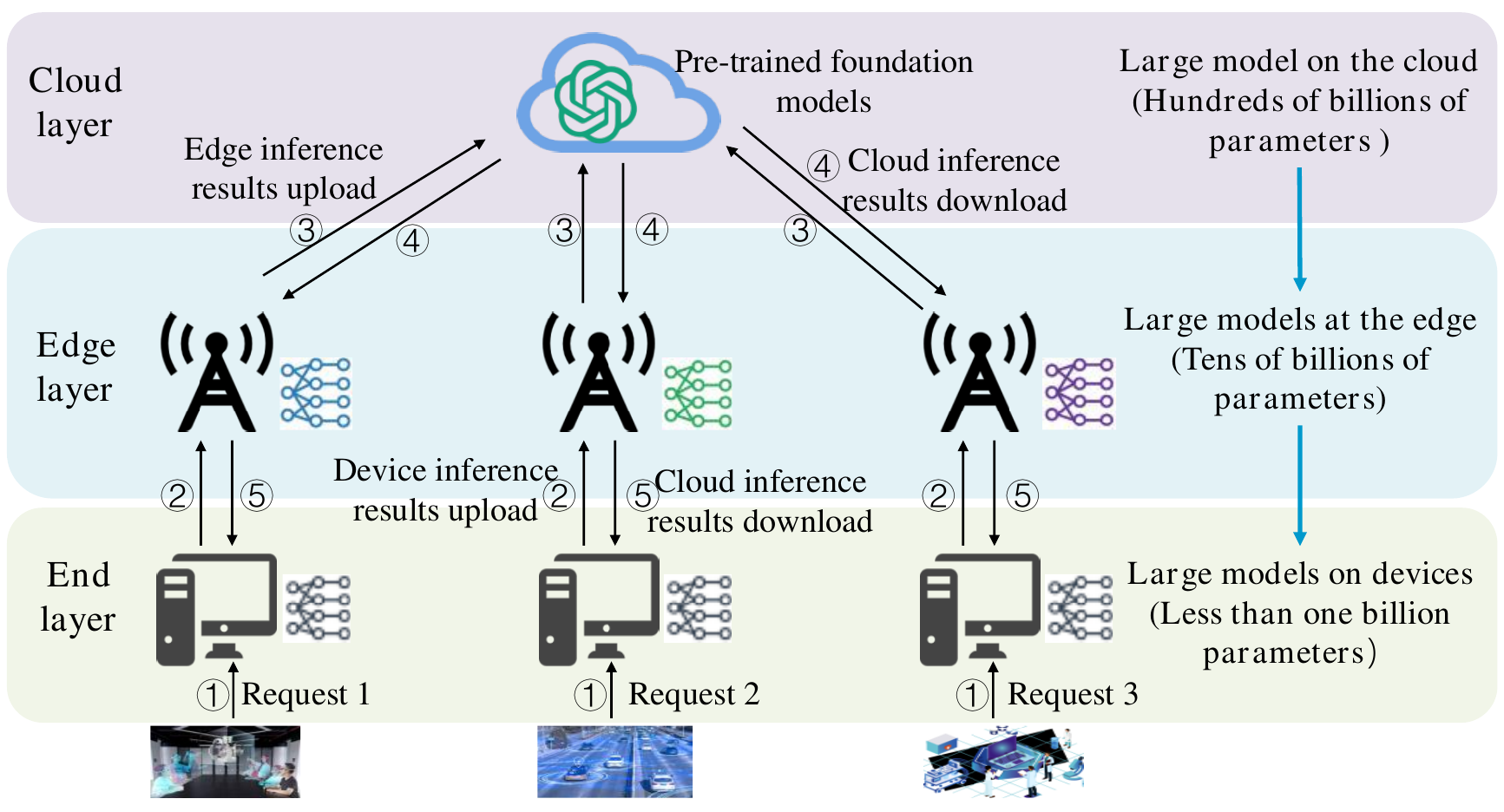}
		\caption{An illustration of cloud-edge-end vertical collaborative reasoning.}
		\label{Fig3}
	\end{figure}
	
	\subsection{Cloud-Edge-End Vertical Collaborative Reasoning}
	Cloud-edge-end collaborative reasoning is a vertical paradigm designed to utilize the computational resources and capabilities of different layers for the rapid inference of intelligent tasks \cite{Wu2024Agglomerative}.
	This multi-layered architecture optimizes performance by leveraging the strengths of each layer. Cloud computing provides extensive storage and processing capabilities, edge computing offers low latency and real-time processing, and end devices benefit from their proximity to data sources and user interactions, all contributing to reduced latency and improved responsiveness.
	As shown in Fig. \ref{Fig3}, in this vertical collaborative framework, cloud servers serve as the central hub for large-scale data storage and complex computations.
	Edge servers, positioned between the cloud and end devices, play a crucial role in reducing latency and bandwidth consumption by performing preliminary data processing, filtering, and even some inference tasks.
	This allows for faster responses and reduces the load on the cloud.
	End devices, such as smartphones, sensors, and IoT gadgets, are the entry points for data collection. They capture raw data from the physical world, which is then sent to edge servers or directly to the cloud for further processing. In the context of collaborative reasoning, end devices may also participate in local computations, utilizing lightweight models to perform initial data analysis or feature extraction.

	\begin{figure}[t]
		\centering
		\includegraphics[width=3.5 in]{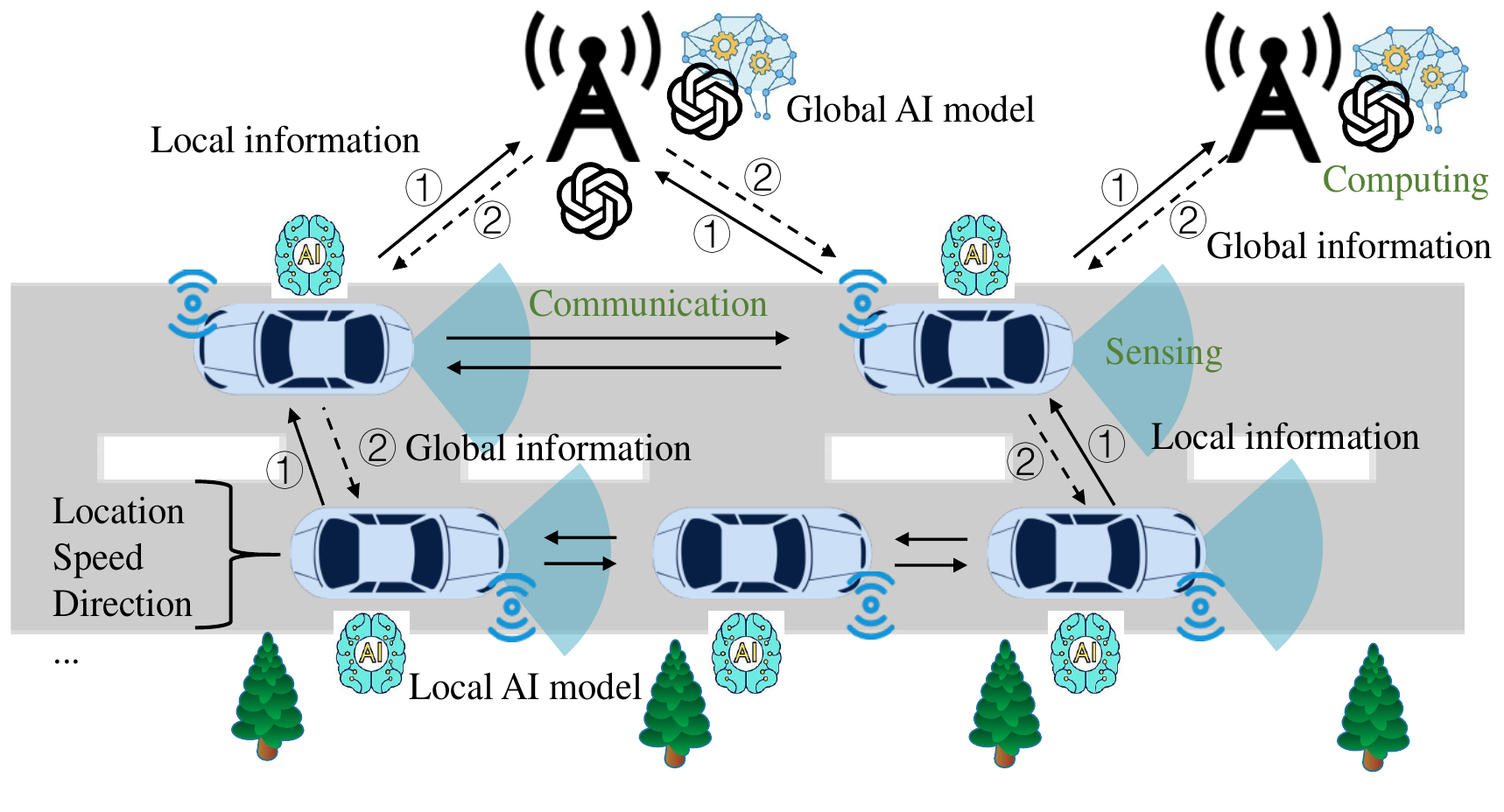}
		\caption{The multi-access collaborative reasoning framework.}
		\label{Fig4}
	\end{figure}

	\subsection{Multi-Access Collaborative Reasoning}
	In multi-access collaborative reasoning framework, a client is capable of directly communicating with multiple edge servers or other end devices, thereby facilitating the offloading of its local computational tasks or inference outcomes to alternative devices and edge nodes. Furthermore, multi-access collaborative reasoning ensures that devices remain interconnected even in the face of network failures or disruptions. This redundancy is critically important for real-time applications, such as autonomous driving, where continuous and reliable communication is essential for maintaining safe and effective operations.
	As illustrated in Fig. \ref{Fig4}, multi-access collaborative reasoning endows vehicles with the flexibility to alternate between various communication pathways, influenced by parameters such as availability, bandwidth, latency, and cost. For instance, a vehicle may utilize cellular networks to connect with a roadside unit (RSU) for disseminating its real-time traffic monitoring data to a broader audience, while concurrently employing vehicle-to-vehicle (V2V) communications for the rapid exchange of real-time inference results among adjacent vehicles. This flexibility ensures that vehicles always have the optimal communication path for reduced latency and improved performance.
	The distributed nature of the multi-access framework makes it more resilient to failures or disruptions. If a RSU experiences downtime, nearby vehicles can continue to process requests, ensuring uninterrupted service.

	\subsection{Resource Allocation for Collaborative Reasoning}
	In wireless networks with distributed devices, the effective implementation of collaborative reasoning for large AI models necessitates a meticulous approach to resource allocation. This is crucial for optimizing performance, reducing latency, and enhancing reliability \cite{Ren2024Industrial}.
	Firstly, the allocation of computational resources plays a pivotal role in collaborative reasoning. Strategies for computational resource allocation include dynamically scaling resources based on workload demands, leveraging cloud and edge computing resources, and optimizing the utilization of available hardware. These strategies ensure that computational tasks are distributed efficiently across the network, thereby minimizing processing time and enhancing overall performance.
	Secondly, the communication overhead and network latency can significantly impact the overall performance of the reasoning process. Therefore, efficient allocation of communication resources is essential to minimize latency and ensure reliable data transfer. This involves optimizing network bandwidth, reducing data redundancy, and implementing efficient data transmission protocols.
	Finally, ensuring synchronization and coordination among distributed nodes is essential for maintaining consistency and coherence in the inference results. This can be achieved through the use of distributed consensus algorithms, which enable nodes to reach agreement on a shared state or decision without requiring a centralized authority. These algorithms ensure that the reasoning process remains coherent and accurate, even in distributed environments.
	
	\begin{figure*} [t]
		\subfloat[]{\label{Fig5_1}
			\begin{minipage}[t]{0.45 \textwidth}
				\centering
				\includegraphics[width= 3.5 in]{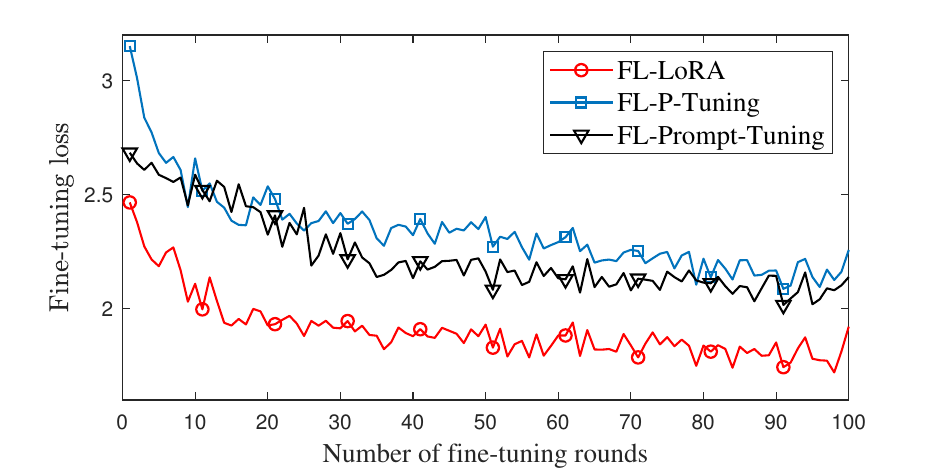}
			\end{minipage}
		} 
		\subfloat[]{\label{Fig5_2}
			\begin{minipage}[t]{0.45 \textwidth}
				\centering
				\includegraphics[width= 3.5 in]{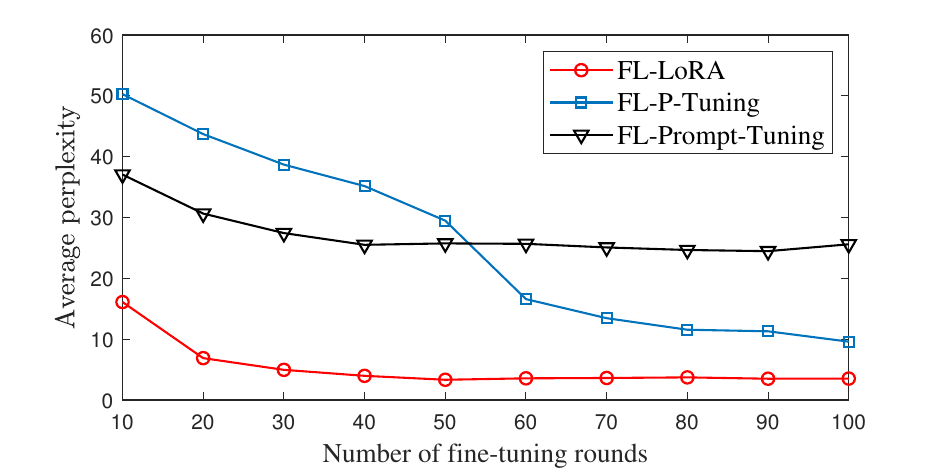}
			\end{minipage}
		}
		\caption{Performance of fine-tuning LLaMA-7B on OpenOrca dataset.}
		\label{Fig5}
		\vspace{-5 mm}
	\end{figure*}

	\begin{figure*} [t]
		\subfloat[Text generation task]{\label{Fig6_1}
			\begin{minipage}[t]{0.45 \textwidth}
				\centering
				\includegraphics[width= 3.5 in]{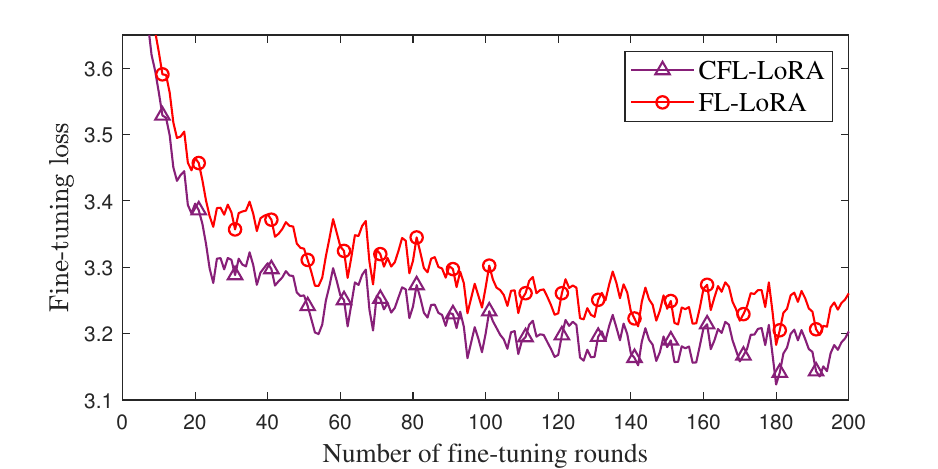}
			\end{minipage}
		}
		\subfloat[Text classification task]{\label{Fig6_2}
			\begin{minipage}[t]{0.45 \textwidth}
				\centering
				\includegraphics[width= 3.5 in]{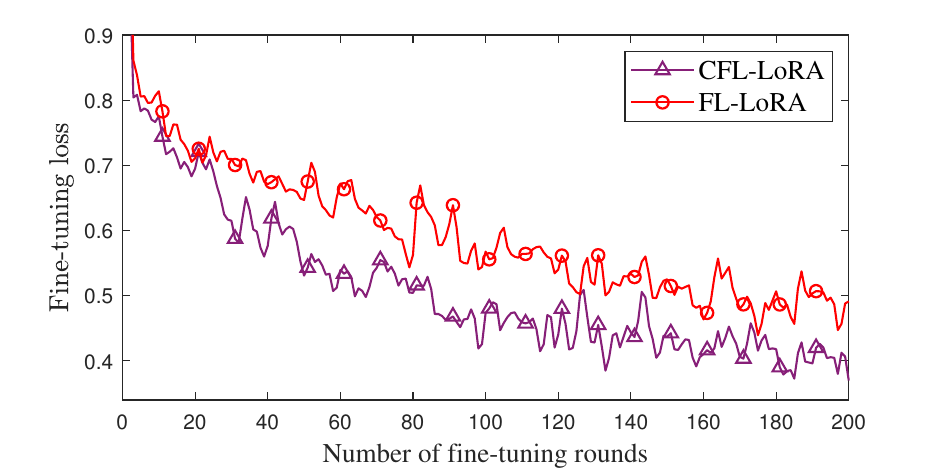}
			\end{minipage}
		}
		\caption{Fine-tuning loss of GPT2-Small on generation and classification tasks.}
		\label{Fig6}
	\end{figure*}

	\section{Simulation Results}
	In this section, we present experimental results to evaluate the performance of fine-tuning large AI models in wireless networks. The setup for our experiments includes a base station with a coverage area of 100 meters in radius and a bandwidth of 20 MHz. We set the number of users to 8 and the number of local samples to 1600, with users deployed randomly within the base station's coverage area. The transmit power of users is set to 23 dBm, while the additive white Gaussian noise is set as -80 dBm.
	For the model training parameters, we use a mini-batch size of 80, a learning rate of $2 \times 10^{-4}$, and a token length of 128.
	To assess performance, we compare the following schemes:
	1) FL-LoRA: A federated low-rank adaptation scheme;
	2) FL-P-Tuning: A federated prefix-tuning scheme;
	3) FL-Prompt-Tuning: A federated prompt-tuning scheme;
	4) CFL-LoRA: A clustered federated fine-tuning scheme.

	Fig. \ref{Fig5} depicts the fine-tuning loss of the LLaMA-7B model on the OpenOrca dataset\footnote{\url{https://huggingface.co/datasets/Open-Orca/OpenOrca}}.
	As shown in Fig. \ref{Fig5_1}, the FL-LoRA scheme exhibits significantly lower fine-tuning loss compared to FL-Prompt-Tuning and FL-P-Tuning. This indicates that FL-LoRA is more effective in adapting the large AI model to the local data distributions.
	In Fig. \ref{Fig5_2}, we further evaluate the fine-tuning performance of the different schemes by calculating the average perplexity.
	The simulation results in Fig. \ref{Fig5_2} indicate a consistent decrease in average perplexity for all three schemes  as the number of fine-tuning rounds increases. However, FL-LoRA achieves the lowest average perplexity, which suggests that the fine-tuned LLM using FL-LoRA is capable of producing more accurate predictions compared to the other methods.

	Fig. \ref{Fig6} shows the fine-tuning performance of GPT2-Small on both text generation and classification tasks using the CFL-LoRA and FL-LoRA schemes. The experiments were conducted with 16 users and 4 clusters.
	The generation task was evaluated using the title-abstract dataset\footnote{\url{https://huggingface.co/datasets/shf123/MDAAG}} spanning four academic disciplines.
	The classification task was evaluated on subsets SST2, QNLI, and QQP from GLUE\footnote{\url{https://huggingface.co/datasets/nyu-mll/glue}}, as well as the Amazon Polarity dataset.
	From Fig. \ref{Fig6_1}, we observe that
	the CFL-LoRA framework achieves faster convergence and lower fine-tuning loss compared to FL-LoRA on the text generation task.
	This improved performance can be attributed to the adoption of a clustered model aggregation approach, which enables CFL-LoRA to dynamically adapt diverse low-rank adapters tailored to the specific domains of multiple clusters.
	Fig. \ref{Fig6_2} illustrates the fine-tuning performance of GPT2-Small on the text classification task.
	Similarly, CFL-LoRA still outperforms FL-LoRA. This further confirms the superiority of the clustered fine-tuning paradigms in enabling more tailored model customization.
	Given that model updates can inadvertently disclose sensitive information about the training data, federated fine-tuning paradigms still pose potential risks of data leakage. Consequently, future research should focus on developing robust aggregation and privacy-preserving techniques to mitigate these risks effectively.
	
	\section{Challenges and Research Opportunities}
	\subsection{Scalability and Computational Efficiency}
	As the size and complexity of AI models increase, the computational resources required for training and inference also grow exponentially. This can lead to bottlenecks in wireless networks, where limited computational power and energy resources must be shared among numerous devices and services. To address this challenge, research is needed to develop scalable AI architectures that can efficiently utilize available resources while maintaining high performance. This includes exploring distributed computing paradigms to offload computational tasks from resource-constrained devices to more powerful servers. Additionally, research into efficient algorithms and hardware accelerators, such as specialized AI chips, can help to alleviate the computational burden of large AI models in wireless networks.
	
	\subsection{Interoperability and Standardization}
	The integration of large AI models into wireless networks also requires addressing interoperability and standardization issues. Different devices, platforms, and services may use different AI models, data formats, and communication protocols, which can lead to compatibility problems and hinder the widespread adoption of AI-based solutions. To address this challenge, research is needed to develop standardized frameworks and protocols for AI-native networks. This includes establishing common data formats, model interfaces, and communication protocols to ensure that different devices and services can seamlessly interact and collaborate.
	
	\subsection{Multi-Modal and Cross-Device Collaboration}	
	Multi-modal large language models (MLLMs) have emerged as powerful tools for processing and understanding diverse types of data. However, their deployment and utilization in real-world applications are hindered by substantial computational resource requirements for both training and inference.
	In wireless networks, data and computational resources are distributed across multiple devices.
	Improved multi-modal fusion techniques can enable MLLMs to better leverage diverse data sources, leading to more accurate and robust models. 
	Efficient cross-device coordination mechanisms ensure that large models can be trained and deployed in scalable and efficient ways, even in resource-constrained IoT networks.
	
	\subsection{Privacy Protection for Opt-Out Users in AI Systems}
	Since large AI models require processing data from multiple users for training and learning, effectively reversing the influence of a user's data on the model when they choose to opt out presents a significant issue. Traditional centralized learning methods are inapplicable in this scenario, as user data, once used for training, cannot be directly removed, thereby complicating the process of revoking its impact on the model.
	Consequently, a critical issue arises concerning how to safeguard user data privacy and prevent data residue upon user opt-out.
	To address this, there is a pressing need to research a federated unlearning mechanism specifically designed for user opt-outs, which can automatically eliminate the influence of a user's data on the model upon their exit, ensuring that the model's accuracy and robustness remain unaffected.
	
	\subsection{Token Communications Between Large AI Models}
	Tokens are the fundamental units processed by large AI models.
	They play a crucial role in connecting bits and semantics by compressing raw data into a more computable format.
	Researchers can delve into optimizing the tokenization process to transform complex multi-modal data into tractable segments for large AI models to process easily.  
	In practical applications, it is imperative to ensure that tokens accurately reflect user intent and mitigate the risk of communication errors stemming from semantic ambiguity or misunderstandings.
	Furthermore, in the context of token communications, tokens act as carriers of data, making their security a paramount concern. It is essential to safeguard tokens against tampering, theft, or misuse during both transmission and storage.

	\section{Conclusions}
	In this paper, we introduced the concept of federated intelligence by integrating federated fine-tuning and collaborative reasoning with large AI models. This concept involved enabling multiple nodes to collaboratively update large AI models and produce a collective output in a distributed environment.
	First of all, federated fine-tuning methods were presented to collaboratively update large AI models while ensuring the protection of user privacy within wireless networks.
	Then, collaborative reasoning frameworks were developed to allocate the inference workload across multiple devices or servers, which enhanced operational efficiency and improved scalability by deploying multiple AI models on distributed nodes.
	Simulation results for different tasks indicated that the proposed fine-tuning methods were effective in improving performance of large AI models in wireless networks.

	\section*{Acknowledgement}
	The work of Wanli Ni was supported in part by the Postdoctoral Fellowship Program of CPSF under Grant Number GZB20240386, and in part by the China Postdoctoral Science Foundation under Grant Number 2024M761669.

	\bibliographystyle{IEEEtran}
	\bibliography{IEEEabrv, refR1}

	\begin{IEEEbiographynophoto}{Wanli Ni} (Member, IEEE) (niwanli@tsinghua.edu.cn) received the B.Eng. and Ph.D. degrees in the School of Information and Communication Engineering from the Beijing University of Posts and Telecommunications (BUPT), China, in 2018 and 2023, respectively. From 2022 to 2023, he was a visiting Ph.D. student at the Nanyang Technological University, Singapore. He is currently a Postdoctoral Assistant Researcher with the Department of Electronic Engineering, Tsinghua University, China. His research interests include federated learning, reconfigurable intelligent surface, semantic communication, edge intelligence, and large AI model. He was a recipient of the Outstanding Doctoral Dissertation Awards from the China Education Society of Electronics in 2023, and also a recipient of the Excellent Graduate of Beijing in 2023. During his doctoral studies, he received the National Scholarship twice, and was recognized as an IEEE Exemplary Reviewer four times.
	\end{IEEEbiographynophoto}
	
	\begin{IEEEbiographynophoto}{Haofeng Sun} (sunhaofeng@bupt.edu.cn) received the B.Eng. degree in communication engineering from Beijing University of Posts and Telecommunications, Beijing, China, in 2023, where he is currently pursuing the Ph.D. degree with the State Key Laboratory of Networking and Switching Technology. His research interests include federated learning, large models, parameter-efficient fine-tuning, over-the-air computation, and wireless resource management.
	\end{IEEEbiographynophoto}
	
	\begin{IEEEbiographynophoto}{Huiqing Ao} (hqao@bupt.edu.cn) received the B.Eng. degree from the School of Information and Communication Engineering, Beijing University of Posts and Telecommunications, China, in 2023, where he is currently pursuing the Ph.D. degree with the State Key Laboratory of Networking and Switching Technology. His research interests include federated learning, split learning, edge intelligence, and wireless resource management.
	\end{IEEEbiographynophoto}
	
	\begin{IEEEbiographynophoto}{Hui Tian} (Senior Member, IEEE) (tianhui@bupt.edu.cn) received the M.S. and Ph.D. degrees from Beijing University of Posts and Telecommunications, China, in 1992 and 2003, respectively. Currently, she is a Professor with the School of Information and Communication Engineering, BUPT. Her research interests include radio resource management in 5G/6G networks, mobile edge computing, cooperative communication, mobile social networks, and Internet of Things.
	\end{IEEEbiographynophoto}

\end{document}